%% file: main.tex
\documentclass{article}

\usepackage{microtype}
\usepackage{graphicx}
\usepackage{subfigure}
\usepackage{booktabs} % for professional tables

\usepackage{array}   % for \newcolumntype macro

\usepackage{hyperref}

\usepackage[accepted]{icml2024}

\usepackage{amsmath,amssymb,amsfonts,amsthm,mathtools}

\usepackage[capitalize,noabbrev]{cleveref}

\theoremstyle{plain}
\newtheorem{theorem}{Theorem}[section]
\newtheorem{proposition}[theorem]{Proposition}
\newtheorem{lemma}[theorem]{Lemma}

\theoremstyle{definition}
\newtheorem{definition}[theorem]{Definition}
\newtheorem{assumption}[theorem]{Assumption}
\theoremstyle{remark}
\newtheorem{remark}[theorem]{Remark}
\newtheorem{example}[theorem]{Example}

\usepackage[textsize=tiny]{todonotes}

\input{macros}

\begin{document}

\twocolumn[
\icmltitle{Convergence of Some Convex Message Passing Algorithms to a Fixed Point} 
%\icmltitle{Convergence of Dual Coordinate Descent to Fixed Point} 




\begin{icmlauthorlist}
\icmlauthor{Václav Voráček}{ut}
\icmlauthor{Tomáš Werner}{ctu}

\end{icmlauthorlist}

\icmlaffiliation{ut}{T\"ubingen AI center, University of T\" ubingen}
\icmlaffiliation{ctu}{Dept. of Cybernetics, Faculty of Electrical Engineering, Czech Technical University in Prague}

\icmlcorrespondingauthor{Václav Voráček}{vaclav.voracek@uni-tuebingen.de}

\icmlkeywords{MAP inference in graphical models, dual block-coordinate descent, message passing, max-sum diffusion, large-scale non-differential optimization}

\vskip 0.3in
]



\printAffiliationsAndNotice{}  % leave blank if no need to mention equal contribution

\begin{abstract}
A popular approach to the MAP inference problem in graphical models is to minimize an upper bound obtained from a dual linear programming or Lagrangian relaxation by (block-)coordinate descent. This is also known as convex/convergent message passing; examples are max-sum diffusion and sequential tree-reweighted message passing (TRW-S).
Convergence properties of these methods are currently not fully understood.
They have been proved to converge to the set characterized by local consistency of active constraints, with unknown convergence rate; however, it was not clear if the iterates converge at all (to any point). We prove a stronger result (conjectured before but never proved): the iterates converge to  a fixed point of the method.
Moreover, we show that the algorithm terminates within $\mathcal{O}(1/\varepsilon)$ iterations.
% \vv{En route to the main result,} we introduce a technique proving convergence of the minimization of maxima of affine functions by coordinate descent under a weak assumption and show that some popular coordinate descent algorithms are special cases of this problem which allowed us to resolve a long standing open problem conjectured e.g., in~\cite{Werner-PAMI07}
We first prove this for a version of coordinate descent applied to a general piecewise-affine convex objective. 
Then we show that several convex message passing methods are special cases of this method.
Finally, we show that a slightly different version of coordinate descent can cycle.
\end{abstract}

\section{Introduction}

Maximum aposterior (MAP) inference in undirected graphical models (Markov random fields) \cite{Wainwright08} leads to an NP-hard combinatorial optimization problem, which is also known as discrete energy minimization \cite{Szeliski06,Kappes-study-2015} or valued constraint satisfaction \cite{Meseguer06,Cooper-AI-2010}.
Graphical models find many applications even in various areas of deep learning~\citep{Chen-ICML-2015,Tourani-ECCV-2018,Munda-GCPR-2017}.

A popular approach to MAP inference is the class of methods known as convex (or convergent) message passing. These are in fact versions of \mbox{(block-)}coordinate descent (BCD) applied to a  dual linear programming (LP) or Lagrangian relaxation/decomposition of the problem, which aims to minimize an upper bound on the true optimal value.
The earliest (and arguably the simplest) such method is max-sum diffusion \cite{Kovalevsky-diffusion}, revisited in \cite{Werner-PAMI07,Werner-PAMI-2010}.
Other examples are sequential tree-reweighted message passing (TRW-S) \cite{Kolmogorov06}, max-product linear programming (MPLP) \cite{Globerson08}, max-marginal averaging \cite{Johnson-allerton2007}, and sequential reweighted message passing (SRMP) \cite{Kolmogorov-PAMI-2015}. 
Message-passing methods for MAP inference have a rich history, reviewing which is beyond the scope of this paper, see also \cite{Meltzer-UAI-2009,Ruozzi-2013,Sontag-MITbook-2012,Tourani-AISTAT-2020}.

% Besides linear programming (LP) relaxation, more powerful convex relaxations can be obtained by Lagrangian decomposition \cite{Guignard-MathProg-1987}. For MAP inference, this was proposed by \cite{Komodakis-PAMI-2011}, see also the review \cite{Sontag-MITbook-2012}.

Besides MAP inference, convex message passing methods have been applied to other combinatorial optimization problems \cite{Wedelin-AOR-1995,Swoboda-CVPR-2017,Swoboda-CVPR-2017b,Swoboda-CVPR-2017c} including the general 0-1 ILPs \cite{Lange-ICML-2021,Abbas-CVPR-2022}, outperforming commercial ILP solvers on many large-scale instances. 

%Dual LP/Lagrangian relaxations of MAP inference are problems with convex piecewise-affine (hence non-smooth) objective function and either no constraints or simple linear equality constraints.
The dual LP or Lagrangian relaxations of MAP inference are convex non-smooth and/or constrained problems.
Coordinate descent applied to a convex function is known to converge to a global minimum if the function is smooth (or its non-smooth part is separable) and has unique coordinate-wise minimizers \cite{Bertsekas99,Tseng:2001} but for non-smooth and/or constrained problems it can get stuck in local (w.r.t.\ coordinate moves) minima \cite{Warga-1963}.
%Indeed, convex message-passing algorithms are known to converge only to local (w.r.t.\ coordinate moves) minima, 
% which are nevertheless often good in practice.
To alleviate this drawback, a number of \emph{globally} optimal large-scale methods have been adapted to solve the relaxations, such as subgradient methods \cite{Komodakis-PAMI-2011}, bundle methods \cite{Savchynskyy-CVPR-2012}, ADMM \cite{Martins:ICL:2011}, or adaptive diminishing smoothing \cite{SavchynskyySKS12}.
It was however observed in the experimental study by \cite{Kappes-study-2015} that for large sparse instances from computer vision, dual BCD methods with tree-structured blocks (such as TRW-S) are consistently faster and the obtained local optima are usually very good.
%In other words, while BCD methods do not guarantee finding global optima of the relaxation, in case of MAP inference this is often more than balanced by their speed and simplicity.

Convergence properties of the convex message-passing methods are currently not fully understood. The objective value cannot increase in any iteration by definition, but many iterations actually keep it unchanged.
It is known that a necessary (but not sufficient) condition for fixed points of the methods is a local consistency (arc consistency for max-sum diffusion, weak tree consistency for TRW-S) of the active constraints.
For TRW-S, \cite{Kolmogorov06} showed that any limit point of the sequence of the iterates satisfies weak tree consistency.
For max-sum diffusion, \cite{Schlesinger-2011} showed somewhat more: the iterates converge to the set defined by arc consistency (but not necessarily to any single point); this was reviewed by \cite{Bogdan-book-2019}. 
This result was generalized by \cite{Werner-CVPR20} to BCD applied to any convex optimization problem, assuming that the block-wise minimizer is always chosen from the relative interior of the set of block-minimizers.

\subsection{Contributions}

As our \emph{first contribution} in this paper, we prove the long-open conjecture, formulated for max-sum difussion by \cite{Schlesinger-2011,Werner-PAMI07} but dating back to \cite{Kovalevsky-diffusion}, that the iterates converge to a fixed point of the method. Moreover, we show that for any precision $\varepsilon>0$ this happens in $\mathcal{O}(1/\varepsilon)$ iterations -- to the best of our knowledge, this is the first result on convergence rate for these methods.
%\tw{Tady to `accuracy' mame zas, c s tom tady? Jak nazvat $\varepsilon$. Je to neco jako 'distance to the set of fixed points' ale ne uplne. Nebo nejak nepresne, treba 'terminating presicion' nebo 'precision parameter in the stopping condition'?}

To that end, we first study (in~\S\ref{sec:aff}) a seemingly simple iterative method, coordinate descent applied to minimization of the pointwise maximum of affine functions.
This method was studied already in \cite{Werner-TR-2017-05}.
%\footnote{It would be possible to extend the theory presented in this section to convex piecewise-affine functions in the form of the \emph{sum} of maxima of affine functions. We restrict ourselves to the form~\refeq{fun} mainly for simplicity.}
Here, the objective function is non-smooth and can have non-unique coordinate-wise minimizers. Therefore, one must decide in each iteration which minimizer to choose. A natural way to resolve this ambiguity is to ignore those affine functions that do not depend on the current variable.

Then (in~\S\ref{sec:convergence}), we introduce a novel energy function that strictly decreases with any non-trivial update. It follows that the method cannot cycle. Assuming boundedness of this energy, we proceed to prove convergence to the fixed point and obtain an assymptotic upper bound on convergence rate.

Finally we show (in~\S\ref{sec:MAP}) that max-sum diffusion and max-marginal averaging in Lagrangian decomposition are special cases of the above algorithm, which allows us to transfer the above convergence results to them. Note that the result for max-marginal averaging is very general, since Lagrangian decomposition is applicable to many hard combinatorial optimization problems beyond MAP inference.

In our \emph{second contribution}, we consider a slight modification of the above method, in which in every update the coordinate-wise minimizer is chosen to be the mid-point of the interval of all coordinate-wise minimizers. We show that in this case, the method can cycle.

\section{Minimizing Maximum of Affine Functions}
\label{sec:aff}

In this section, we consider coordinate descent applied to unconstrained minimization of a convex piecewise-affine function. Such a function can be always expressed as the pointwise maximum of affine functions,
\begin{equation}
f(x) = \max_{i\in[m]}(a_i^Tx+b_i)
= \max_{i\in[m]} (Ax+b)_i ,
\label{eq:fun}
\end{equation}
where $a_i^T$ are the rows of matrix $A=[a_{ij}]\in\mathbb R^{m\times n}$ and $b=(b_1\...b_m)\in\mathbb R^m$. We aim at applications where $A$ is large, sparse and its non-zero entries are small integers, often just $\{-1,0,+1\}$. We will refer to $a_{i1},\dots,a_{in}$ as the \emph{coefficients}
%\tw{What about calling $a_{ij}$ `slopes' instead of 'coefficients'? If so, please change it everywhere in the paper.} 
and to~$b_i$ as the \emph{offset} of the $i$th affine function $a_i^Tx+b_i$.

Starting from an initial point $x=(x_1\...x_n)\in\mathbb R^n$, in each iteration of coordinate descent we pick some $j\in[n]$ and minimize~$f$ over variable~$x_j$ while keeping the remaining variables $x_1,\dots,x_{j-1},x_{j+1},\dots,x_n$ fixed, i.e., we minimize the univariate function $x_j\mapsto f(x)$.

Here we assume that that the function $x_j\mapsto f(x)$ always has a minimum. This is ensured by the following condition, which we assume throughout the paper:
\begin{assumption}\label{technical_assumption}    
The matrix $A$ satisfies
\begin{equation}
\forall j: (( \exists i: a_{ij}<0 )\, {\wedge}\, ( \exists i: a_{ij}>0 ) ) .
\label{eq:N-scons}
\end{equation}
\end{assumption}
It says that for each variable~$x_j$ there is at least one affine function increasing in~$x_j$ and at least one affine function decreasing in~$x_j$.
If this is not the case for some~$j$, the rows with non-zero elements in column~$j$ can be deleted from~$A$ without affecting the infimum of~$f$, because the corresponding affine functions can be decreased arbitrarily without changing the other affine functions. This can be repeated until \eqref{eq:N-scons}~becomes satisfied (if this makes~$A$ empty, then $f$~is unbounded from below or constant). Thus, assumption~\eqref{eq:N-scons} is purely technical and does not limit the following results in any way.
We remark that \eqref{eq:N-scons}~is a form of local consistency, called \emph{sign consistency} in \cite{Werner-TR-2017-05}.

% We assume throughout the paper that each column of~$A$ contains a negative coefficient and a positive coefficient, i.e.,
% %each column of~$A$ contains a negative element and a positive element,
% \begin{equation}
% \forall j\colon (( \exists i\colon a_{ij}<0 )\, {\wedge}\, ( \exists i\colon a_{ij}>0 ) ) .
% \label{eq:N-scons}
% \end{equation}
% That is, for each variable~$x_j$ there is at least one affine function increasing in~$x_j$ and at least one affine function decreasing in~$x_j$.
% If this is not the case for some~$j$, the rows with non-zero elements in column~$j$ can be deleted from~$A$ without affecting the infimum of~$f$, because the corresponding affine functions can be decreased arbitrarily without changing the other affine functions. Similarly, any all-zero column can be deleted. This can be repeated until \eqref{eq:N-scons}~becomes satisfied or $A$~becomes empty (in the latter case, $f$~is unbounded below or constant). Thus, assumption~\eqref{eq:N-scons} is purely technical and does not limit the following results in any way.
% We remark that \eqref{eq:N-scons}~is a form of local consistency, called \emph{sign consistency} in \cite{Werner-TR-2017-05}.

\medskip
\begin{example}
The function $f(x_1,x_2)=\max\{x_1,\,-x_1$, $x_1+x_2\}$ does not satisfy~\eqref{eq:N-scons} because of variable~$x_2$. But the affine function $x_1+x_2$ can be omitted  without affecting the minimum of~$f$. The new function $f(x_1,x_2)=\max\{x_1,\,-x_1\}$ satisfies~\eqref{eq:N-scons}.
\end{example}

So we have ensured that the univariate function $x_j\mapsto f(x)$ always has at least one minimizer. But it may have more than one minimizer (an interval) because some of the affine function can be constant in~$x_j$. Therefore, we need to introduce a rule to choose a single minimizer. A~natural\footnote{Another natural rule, see is to choose a mid-point of the interval of minimizers. We show in~\S\ref{sec:midpoint} that such a rule may lead to oscillation of iterates.}  such rule, considered in \cite{Werner-TR-2017-05}, is to \emph{ignore} the affine functions that are constant in~$x_j$, i.e., to replace~$f$ with the function
\begin{equation}
g_j(x) = \max_{i:\,a_{ij}\neq0} (a_i^Tx+b_i) .
\label{eq:phi}
\end{equation}

Assuming~\refeq{N-scons}, for every $x_1\...x_{j-1},x_{j+1}\...x_n\in\mathbb R$ the univariate function $x_j\mapsto g_j(x)$ has exactly one minimizer, $x^*_j$, which is the unique solution to the equation
\begin{equation}
\max_{i:\,a_{ij}<0} (a_i^Tx+b_i) = \max_{i:\,a_{ij}>0} (a_i^Tx+b_i) .
\label{eq:fixedpt-x}
\end{equation}
For the special case $A\in\{-1,0,1\}^{m\times n}$, \eqref{eq:fixedpt-x} has the closed-form solution
\begin{equation}
x^*_j = \frac12\Bigl(\, \max_{i:\,a_{ij}<0} (a_i^Tx^{-j}+b_i) - \max_{i:\,a_{ij}>0} (a_i^Tx^{-j} + b_i) \Bigr)
%x^*_j = \frac12\Bigl( \max_{i:\,a_{ij}<0} \bigl({\textstyle\sum\limits_{k\neq j}} a_{ik}x_k+b_i \bigr) - \max_{i:\,a_{ij}>0} \bigl({\textstyle\sum\limits_{k\neq j}} a_{ik}x_k + b_i \bigr) \Bigr) .
\label{eq:fixedpt-x-101}
\end{equation}
where $x_k^{-j} = x_k$ for all $k \neq j$ and $x_j^{-j}=0$.
%The unique minimizer of the function $x_j\mapsto g_j(x)$ is always a relative interior point of the minimal set (an interval or a singleton) of the function $x_j\mapsto f(x)$, so the rule complies to \cite{Werner-CVPR20}.

\begin{figure}[t]
\centering
\includegraphics[width=.9\linewidth]{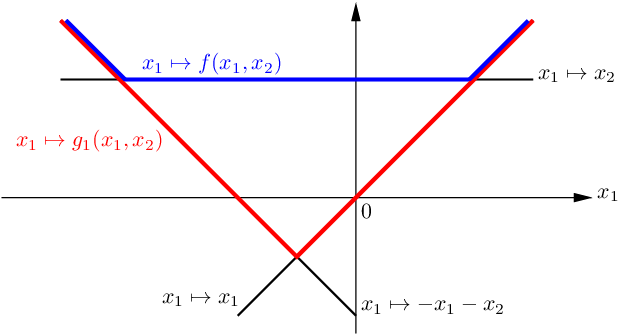}
\vskip-2ex
\caption{Plots of the functions $x_j\mapsto f(x)$ (in blue) and $x_j\mapsto g_j(x)$ (red) for the first update in Example~\ref{ex:nonunique} (so that $x_2=1$). Also shown are the three constituent affine functions (black).}
\label{fig:max_aff}
\end{figure}

% \begin{figure}[t]
% \centering
% \includegraphics[width=.9\linewidth]{maxaff}
% \vskip-3ex
% \caption{Example of (non-)uniqueness of a minimizer of the \strike{maxima of} functions \eqref{eq:fun} and \eqref{eq:phi} in red and blue, respectively.
% \tw{Misto $x$ ma v obrazku byt $x_j$ a misto $g$ ma byt $g_j$.}
% }
% \label{fig:max_aff}
% \end{figure}

\medskip
\begin{example}\label{ex:nonunique}
Let
\begin{equation}
f(x_1,x_2) = \max\{ x_1, \, x_2, \, -x_1-x_2 \}.
\label{eq:nonunique}
\end{equation}
According to~\eqref{eq:phi}, we thus have
\begin{align*}
g_1(x_1,x_2) &= \max\{ x_1, \, -x_1-x_2 \} ,\\
g_2(x_1,x_2) &= \max\{ x_2, \, -x_1-x_2 \} .
\end{align*}
Let us show two iterations of the method, starting from the point $(x_1,x_2)=(1,1)$.

To update~$x_1$, we minimize the univariate function $x_1\mapsto g_1(x_1,x_2)=\max\{x_1,-x_1-1\}$, which has the unique minimizer $x_1=-\frac12$. In contrast, the function $x_1\mapsto f(x_1,x_2)=\max\{x_1,1,-x_1-1\}$ attains its minima on the interval $[-2,1]$, which contains the point~$-\frac12$ (see Figure~\ref{fig:max_aff}).
Note that this update actually did not decrease the objective~\eqref{eq:nonunique}, because $f(1,1)=f(-\frac12,1)=1$. 

To update~$x_2$, we minimize the function $x_2\mapsto g_2(x_1,x_2) = \max\{x_2,\, \tfrac12-x_2\}$, which has the unique minimizer $x_2=\frac14$. In this case, the function $x_2\mapsto f(x_1,x_2) = \max\{-\frac12,\, x_2,\, \tfrac12-x_2\}$ has the same unique minimizer. This update decreased the objective: $f(-\frac12,1)=1 > f(-\frac12,\frac14)=\frac14$.

One might think that the updates that do not decrease the objective can be skipped -- but they cannot. Indeed, $f$~cannot be decreased from the point $(x_1,x_2)=(1,1)$ by changing any variable separately but the method nevertheless converges to the global minimizer $(x_1,x_2)=(0,0)$ of~$f$.
\end{example}

% \begin{example}\label{ex:nonunique}
% Let
% \[
% f(x_1,x_2,x_3) = \max\{ x_2-x_3, \, x_3-x_1, \, x_1-x_2 \}.
% \]
% Suppose $(x_1,x_2,x_3)=(0,1,0)$. To minimize~$f$ over~$x_1$, we need to minimize the univariate function $x_1\mapsto f(x_1,x_2,x_3)=\max\{1,-x_1,x_1-1\}$, which attains its minimal value~1 on the interval $[-1,2]$.
% However, for
% \[
% g_1(x_1,x_2,x_3) = \max\{ x_3-x_1, \, x_1-x_2 \},
% \]
% the function $x_1\mapsto g_1(x_1,x_2,x_3)=\max\{-x_1,x_1-1\}$ has the single minimizer $x_1=\frac12$.
% \end{example}

%Thus, in each iteration of the method we pick some $j\in[n]$ and adjust variable~$x_j$ to satisfy~\refeq{fixedpt-x}, keeping the other variables unchanged.
The above iterative method is summarized in Algorithm~\ref{alg:maxaff}, including a stopping condition.
A {\em fixed point\/} of the method is any point that satisfies~\refeq{fixedpt-x} for all~$j\in[n]$.
%Note, such a point is a Nash equilibrium for penalty functions $g_1\...g_n{:}\ \mathbb R^n\to\mathbb R$.

\begin{algorithm}
\caption{Minimizing pointwise maximum of affine functions by coordinate descent}\label{alg:maxaff}
\begin{algorithmic}
\STATE {\bf Input:} data $A,b$,\, initial point $x\in\mathbb R^n$, precision $\varepsilon \ge 0$
\STATE $\eta \gets \infty$
\WHILE{$\eta \geq \varepsilon$}
\STATE $\eta \gets 0$
\FOR{$j\in[n]$}
\STATE $x^*_j \gets \argmin\limits_{x_j\in\mathbb R} g_j(x)$
\STATE $\eta \gets \max\{ |x_j-x_j^*|, \eta \}$
\STATE $x_j\gets x_j^*$
\ENDFOR
\ENDWHILE
\end{algorithmic}
\end{algorithm}

% Moreover, versions with and without messages can be conveniently described as 'updating variables $x$' and 'updating offsets $b$' (that's one reason why I introduced these words in the beg of this section).

% (BTW, if we consider version with changing offests, I think that it would be better to denote offsets by $y$ than by $b$.)
% }

\begin{remark}\label{rem:msgfree}
One can consider an equivalent version of the algorithm, where we do not explicitly keep the variables~$x$ in the memory but instead modify the offsets~$b$.
Indeed, the vector $Ax+b$ does not change if we set $b\gets Ax+b$ and then reset $x\gets 0$. Detail can be found in \cite{Werner-TR-2017-05}.
%In fact, this might be a natural way to implement the algorithm since the naive implementation as in Algorithm~\ref{alg:maxaff} would require a matrix-vector multiplication in every iteration \tw{I suggest to omit the previous sentence -- in fact, this is not done in MAP inference because the number of $b$'s can be much larger than the number of $x$'s)}.
In MAP inference literature, this corresponds to `message-free' versions of message-passing algorithms.
\end{remark}

Of course, the method still may have fixed points that are not global minima. For example, the function
\begin{equation}
f(x_1,x_2)=\max\{x_1-2x_2,\,x_2-2x_1\}
\end{equation}
is unbounded but any point $x_1=x_2$ is fixed for the method. This is a well-known drawback of coordinate descent applied to non-smooth convex functions \cite{Warga-1963}.

Next we present an example when some components of the vector $Ax+b$ (i.e., the values of some affine functions) decrease unboundedly during the iterations.

\medskip
\begin{example}[\cite{Werner-TR-2017-05}]
\label{ex:divergence}
Let
\begin{equation}\label{eq:example_divergence}
\begin{split}
\hskip-2ex
f(x_1, x_2,x_3) = \max\{ & x_1-x_2-x_3, \, x_1+4, \\ & x_1+x_2+x_3, \, -x_1+x_2+2 \} .
\end{split}
\end{equation}
Starting from $(x_1,x_2,x_3) = (0,0,0)$, the updates of $x_1,x_2,x_3$ (in this order) read
\begin{align*}
x_1 \gets -1 &= \argmin_{x_1} \max\{x_1, \, x_1+4, \, x_1, \, -x_1+2\},\\
x_2 \gets -2 &= \argmin_{x_2} \max\{-x_2-1, \, x_2-1, \, x_2+3\}, \\
x_3 \gets \phantom{-}2 &= \argmin_{x_3} \max\{1-x_3, \, -3+x_3\}.
\end{align*}
By these three updates, each component of the vector $Ax+b$ decreases by~1. This can be seen as decreasing each component of~$b$ by~1 and resetting $(x_1,x_2,x_3)\gets(0,0,0)$, as in Remark~\ref{rem:msgfree}. Thus, the vector $Ax+b$ will diverge.
\end{example}

% \begin{example} [\citep{Werner-TR-2017-05}]
% \label{ex:divergence}
% Let
% \begin{equation}\label{eq:example_divergence}
%     f(x, y, z) = \max\{\, x-y-z, \, x+4, \, x+y+z, \, -x+y+2 \,\}
% \end{equation}
% (where we use $(x,y,z)$ instead of $(x_1, x_2, x_3)$ for the sake of clarity).
% We show that Algorithm~\ref{alg:maxaff} diverges on this example when we start with $(x,y,z) = (0,0,0)$. The first three iterations in $x,y,z$ in this order are as follows:
% \begin{align*}
% x^* = -1 &= \argmin_x \max\{x, \, x+4, \, x, \, -x+2\},\\
% y^* = -2 &= \argmin_y \max\{-y-1, \, y-1, \, y+3\}, \\
% z^* = 2 &= \argmin_z \max\{1-z, \, -3+z\}.
% \end{align*}
% Thus, we decreased every affine function from Equation~\eqref{eq:example_divergence} by $1$. In the view of the ``message-free" notation, we decreased vector $b$ by $1$ and returned to the initial position; thus, $Ax+b$ will diverge.
% \end{example}

It might seem this behaviour cannot occur if the objective~$f$ is bounded below -- but it is not so. Consider, for example, the function $f'(x_1,x_2,x_3)=\max\{f(x_1,x_2,x_3),0\}$ where $f$~is given by~\refeq{example_divergence}, which is bounded below by~0. The zero affine function will be ignored in all iterations (not involved in any function~$g_j$), so the values of the remaining four affine functions will be decreasing exactly the same way as in Example~\ref{ex:divergence}.

\subsection{Convergence Analysis}
\label{sec:convergence}

The key ingredient of our convergence analysis is a novel `energy function' \refeq{lex-energy} that stricly decreases in every non-trivial (i.e., that changes some variable) update. Existence of such a function is not at all obvious -- recall, in particular, that the objective $f(x)$ can remain unchanged in some iteration and these iterations cannot be omitted, see Example~\ref{ex:nonunique}.

% To analyze the convergence properties of the method, instead on vectors~$x$ we will focus on vectors $Ax+b$ prove. This makes sense because the atrix~$A$ can have a non-trivial nullspace and the vectors $Ax+b$ have a clearer interpretation in MAP inference tah vecttors~$x$, cf.\ Remark~\ref{rem:msgfree}. First we show that Algorithm~\ref{alg:maxaff} cannot oscillate. To do so, we define an energy that is strictly decreased in every non-trivial iteration. Thus, the algorithm cannot return to its previous state. This is not sufficient for the convergence as it turns out that for general problems of minimization of the maxima of affine functions, the vector $Ax+b$ may actually diverge, as shown in the following example from \citep{Werner-TR-2017-05}. 
 
% However, in many cases, such as in the max-marginal averaging or in the max-sum diffusion, the structure of the problem is pleasant enough that we are able to prove that they converge. In particular, when for an algorithm we can show that every entry of the vector $Ax+b$ is bounded from below, we can show that the energy is lower-bounded and the algorithm enjoys convergence in $\mathcal{O}(1/\varepsilon)$ steps.

\begin{definition}\label{def:energy}
For any $y\in\mathbb R^m$ and $k \geq 2$, define
\begin{subequations}
\begin{align}
E_k(y, \pi) &= \sum_{i\in[m]} k^i\, y_{\pi(i)}, \\
E_k(y) &= \max_\pi E_k(y, \pi) = \sum_{i\in[m]} k^i\sort(y)_i
\label{eq:lex-energy}
\end{align}
\end{subequations}
where $k^i$ denotes the $i$th power of~$k$, and in~\refeq{lex-energy} we maximize over all permutations~$\pi$ and the sort is in ascending order.
%\vv{ale tohle neděláme, ne?}For brevity, we will sometimes omit the subscript when $k=2$, so $E = E_2$.
\end{definition}

Note that the function~\eqref{eq:lex-energy} is convex (although we will not need this property).
While we present the function in the full generality, for MAP inference (in~\S\ref{sec:MAP}) it will suffice to consider only the case $k=2$.

Each iteration of the method minimizes, for some~$j$, the univariate function $x_j\mapsto g_j(x)$, which is the pointwise maximum of univariate affine functions. This decreases some of the affine functions and increases some other. We have a bound for this increase because we know both the minimal and maximal slope 
\begin{equation}
c = \min_{i,j:\,a_{ij}\neq 0} \abs{a_{ij}}, \qquad
C = \max_{i,j} \abs{a_{ij}}
\label{eq:slopes}
\end{equation}
of the affine functions. This allows us to prove that the energy indeed stricty decreases.

\begin{proposition}\label{prop:energy_decrease}
Let Assumption~\ref{technical_assumption} hold.
Let $c$ and~$C$ be given by~\refeq{slopes}.
In every  {inner iteration} of Algorithm~\ref{alg:maxaff}, the energy $E_{1+C/c}(Ax+b)$ decreases by at least $c\,|x_j - x_j^*|$.
\end{proposition}

\vspace*{-2ex}
\begin{proof}
Let $x^*_j$ be the minimizer of the function $x_j\mapsto g_j(x)$. There is an affine function $a_i^Tx +b_i$ with a positive slope and another with a negative slope in~$x_j$, which at~$x^*_j$ are equal to the minimal value of the function $x_j\mapsto g_j(x)$ as in Equation~\eqref{eq:fixedpt-x}. 
% There is an affine function with positive and another  with negative slope which equal o~\eqref{eq:phi} at $x_{j}^*$ as otherwise $0 \not \in \partial g_j(x_{j}^*)$, contradicting the fact that $x_{j}^*$ is the minimizer of $g_j$ which is guaranteed to exist due to Assumption~\ref{technical_assumption}.
Thus, there is an affine function for which the minimum  of the function $x_j\mapsto g_j(x)$ is attained at $x_j^*$ whose value decreased by at least $c\,|x_j - x_j^*|$ in this iteration. The values of all the other affine functions could not increase by more than $C\,|x_j - x_j^*|$. By Lemma~\ref{lem:energy_2}, the energy $E_{1+C/c}(Ax+b)$ decreased by at least $c\,|x_j - x_j^*|$ in this iteration.
\end{proof}

%\tw{Nasledujic lemma neoptrebuje predpokladat \refeq{N-scons}, souhlasite?}
% jojo

\begin{lemma}\label{lem:energy_2}
Let $c,C>0$.
Let $y,y' \in \R^m$ differ at positions $I\subseteq[m]$ satisfying $y'_i \leq y_i+C$ for all $i\in I$, and at the same time, there is also a position $i^* \in \Argmax_{i\in I} y'_i$ such that $y'_{i^*} \leq y_{i^*} - c$.
Then  
\[
E_{1+C/c}(y) \geq E_{1+C/c}(y')+ c.
\]
\end{lemma}

\vspace*{-2ex}
\begin{proof}
Let  $\pi$ be a permutation sorting~$y'$ 
%\tw{Zde ma byt $y'$, i ve footnote} 
in an increasing order\footnote{That is, $y'_{\pi(1)} \leq y'_{\pi(2)} \leq \dots \leq y'_{\pi(m)}$. The inverse $\pi^{-1}$ is then mapping original indices to their order.} such that $\pi^{-1}(i^*) \geq \pi^{-1}(i)$ for all $i \in I$. Let $j = \pi^{-1}(i^*)$. Thus, in particular, we have $y_{\pi(i)} = y'_{\pi(i)}$ for $i>j$. Now

\begin{align*}
&E_{1+C/c}(y,\pi) - E_{1+C/c}(y',\pi) \\ 
&= \sum_{i\in[m]} (y-y')_{\pi(i)}\left( 1+C/c\right)^i\\
&\geq c\left(1+C/c\right)^j  - \sum_{i=1}^{j-1} C \left(1+C/c\right)^i\\
&= c\left(1+C/c\right)^j - \frac{C\bigl((1+C/c)^j-1\bigr)}{1+C/c -1}\\
&=  c
\end{align*}
where the first equality follows from the definition of the energy{, using} the assumed relations between $y_i$ and $y'_i$. In particular, we used $y'_i \leq y_i+C$ for the first $j-1$ elements of $(y-y')_{\pi}$, and $y'_{i^*} \leq y_{i^*}-c$ for the $j$th one.  In the second equality we used the formula for the sum of geometric series.
    Finally, omitting the subscripts for brevity, we have 
    \[
    E(y) \geq E(y, \pi) \geq E(y',\pi) +c =  E(y') + c ,
    \]
    which finishes the proof.
\end{proof}

Existence of a strictly decreasing quantity shows that the method cannot get into a cycle. This is not yet enough for convergence because, as we saw in Example~\ref{ex:divergence}, some components of the vector $Ax+b$ (and hence the energy) can decrease unboundedly. We currently do not know about any concise condition characterizing when this happens. Therefore, in the following theorem, the main result of the paper, we explicitly assume that the iterates are bounded during the algorithm which will be the case for MAP inference in~\S\ref{sec:MAP}.

%\tw{Zmenil jsem trochu zneni tohoto teoremu i analogickych teoremu v sekcich o MAP inference -- check please. V Algoritmech jsem na vstupu dovolil $\varepsilon\ge0$ misto $\varepsilon>0$:}

\begin{theorem}\label{thm:main_aff}
Let Assumption~\ref{technical_assumption} hold.
Let $c$ and $C$ be given by~\refeq{slopes}.
Let the energy $E_{1+C/c}(Ax+b)$ 
%remain 
be bounded during Algorithm~\ref{alg:maxaff}.
For any $\varepsilon>0$, the algorithm halts in $\mathcal{O}(1/\varepsilon)$ iterations.
For $\varepsilon = 0$, we have $\lim_{t \to \infty} \eta^t = 0$ where $\eta^t$~denotes the value of~$\eta$ after $t$~outer iterations, and both $x$ and $Ax+b$~converge. 
\end{theorem}

\vspace*{-2ex}
\begin{proof}
The energy is decreased in every step, and can decrease only by some constant $M$ in total due to the boundedness. Within every outer iteration, there has to be a decrease in energy by at least~$c\varepsilon$ by Proposition~\ref{prop:energy_decrease}; otherwise the algorithm would terminate. Thus, there can be at most $M/(c\varepsilon) = \mathcal{O}(1/\varepsilon)$ non-terminating outer iterations before the energy is at its minimum.

For the convergence of iterates, we reiterate the previous argument. When $x_j$  changes by $\varepsilon$, then the energy decreases by at least~$c\varepsilon$. Thus, boundedness of energy implies the boundedness of $\sum_{t=1}^\infty \|x^t - x^{t+1}\|_\infty$ (where $x^t$ denotes $x$ after $t$~inner iterations). So $x$ cannot diverge nor, by Proposition~\ref{prop:energy_decrease}, can it oscillate between multiple limit points;  convergence of $x$ implies convergence of $Ax+b$.
\end{proof}

We remark that the convergence of~$x$ or of $Ax+b$ does not follow from the fact that Algorithm~\ref{alg:maxaff} halts after at most $\mathcal{O}(1/\varepsilon)$ steps. Clearly, if the sequence~$(\eta_t)$ was, e.g., $(1, \frac12, \frac13, \frac14, \dots)$, then the algorithm would halt after $1/\varepsilon$ steps but the sequences of $x$ and $Ax+b$ would not necessarily converge.

\section{MAP Inference}
\label{sec:MAP}

Now we turn our attention to the combinatorial optimization problem arising in MAP inference. It can be described as follows. We are given a set $V$ of  variables, each taking states from a finite label set $L$, and a set of weight functions, each depending on a (small) subset of the variables. We aim to maximize the sum of the functions over all the variables.
For illustrative examples see, e.g., \cite{Wainwright08,Bogdan-book-2019,Kappes-study-2015}.
For simplicity of presentation, we consider its pairwise version, when the functions can depend only on individual variables or pairs of variables. This problem reads\footnote{In this section, symbol $x$ has a different meaning that in \S\ref{sec:aff}.}
\begin{equation}
F(\theta) = \max_{x\in L^V} \Bigl[ \sum_{i\in V} \theta_{i,x_i} + \sum_{\{i,j\}\in E} \theta_{ij,x_ix_j} \Bigr]
\label{eq:maxsum}
\end{equation}
where $(V,E)$ with $E\subseteq{V\choose2}$ is an undirected
graph,
% $L$~is a label set,
$\theta_{i,x}$ ($i\in V$, $x\in L$) are unary weights, and 
$\theta_{ij,xy}$ ($\{i,j\}\in E$, $x,y\in L$) are binary weights (adopting that
$\theta_{ij,xy}=\theta_{ji,yx}$). All the weights together form a vector $\theta\in\mathbb R^I$ where
\begin{equation}
I = (V\times L) \cup \set{\{(i,x),(j,y)\} \mid \{i,j\}\in E, \; x,y\in L} .
\label{eq:I}
\end{equation}
A labeling $x\in L^V$ assigns label $x_i\in L$ to each variable $i\in V$.

%\tw{Tento priklad bych vyhodil v pripade nedostatku mista:}
% místo je 

\medskip
\begin{example}
Consider an image segmentation problem where we want to find a dark object on a bright background on a grayscale image. The set of variables $V$ corresponds to the image pixels, and two vertices are connected by an edge $e\in E$ if the corresponding pixels are neighbouring. There are two labels in this task $L = \{\text{object}, \text{background}\}$. The unary weights might encode our prior knowledge that the object is dark, and the background is bright in the following way: $\theta_{i,\text{object}} = 1-J(i)$ and $\theta_{i,\text{background}} = J(i)$, where $J(i)$ is the image intensity of pixel $i$. The pairwise weights may encode the fact that both the object and background are continuous, so we should expect neighbouring pixels to share label; then we might want $\theta_{ij,xy} = \mathbf{1}_{x=y}$.
\end{example}

\subsection{Max-Sum Diffusion}

It is well-known (see, e.g., \cite{Werner-PAMI07}) that the objective function of~\refeq{maxsum} can be \emph{reparameterized} by adding constants to some weights and subtracting the same constants from some other weights. The simplest such reparameterization acts on a single triplet $(i,j,x)\in P$, where
\begin{equation}
P=\set{(i,j,x)\mid i\in V, \; j\in N_i, \; x\in L}
\label{eq:P}
\end{equation}
and $N_i=\set{j\in V\mid \{i,j\}\in E}$, as follows:
subtract a number (`message') $\delta_{ij,x}$ from~$\theta_{i,x}$ and add the same number~$\delta_{ij,x}$ to~$\theta_{ij,xy}$ for each $y\in L$. Clearly, this preserves the objective of~\refeq{maxsum} because $\delta_{ij,x}$ cancels out in the sum. Composing these elementary reparameterization for all $(i,j,x)\in P$ changes the initial weight vector $\theta\in\mathbb R^I$ to the vector $\theta^\delta\in\mathbb R^I$ given by
\begin{subequations}
\label{eq:reparam}
\begin{align}
\theta_{i,x}^\delta &= \theta_{i,x} - \sum_{j \in N_i}\delta_{ij,x} \label{eq:reparam:1}\\
\theta_{ij,xy}^\delta &= \theta_{ij.xy} + \delta_{ij,x} + \delta_{ji,y} \label{eq:reparam:2}
\end{align}
\end{subequations}
where all the messages $\delta_{ij,x}$ form a vector $\delta\in\mathbb R^P$.

Many LP-based MAP inference algorithms minimize a convex
piecewise-affine upper bound on~\refeq{maxsum} over
reparameterizations. We consider two such quantities
\begin{align*}
U_1(\theta) &= \sum_{i\in V} \max_{x\in L} \theta_{i,x} + \sum_{\{i,j\}\in E} \max_{x,y\in L} \theta_{ij,xy} , \\
U_2(\theta) &= \max\Bigl\{ \max_{i\in V} \max_{x\in L} \theta_{i,x} , \max_{\{i,j\}\in E} \max_{x,y\in L} \theta_{ij,xy} \Bigr\} ,
\end{align*}
which clearly upper-bound~\refeq{maxsum} as
\begin{equation}
F(\theta) \le U_1(\theta) \le (|V|+|E|)\,U_2(\theta) .
\label{eq:F<U<U}
\end{equation}

To obtain the best upper bound, we want to minimize $U_1(\theta^\delta)$ or $U_2(\theta^\delta)$ over~$\delta$.
This can be formally obtained as a dual LP relaxation of~\refeq{maxsum}. 
If the graph $(V,E)$ is connected, at optimum we have $U_1(\theta^\delta)=(|V|+|E|)U_2(\theta^\delta)$, so these two relaxations are equivalent \cite{Werner-PAMI07}.

% \vv{maybe say that people use CD methods here because of sparsity and different CD methods have different convergence properties (some of them does not converge at all) depending on how do we choose a minimizer from a set of minimizers?}
% \tw{Tohle je obtizne psat, na to nemam silu. Tech metod je hodne a diskuze, ktera je dobra na co, je out of scope of this paper.}

Arguably the simplest convex message passing method to minimize the above upper bound is known as \emph{max-sum diffusion} \cite{Kovalevsky-diffusion,Schlesinger-2011,Werner-PAMI07}. It has been formulated in several slightly different versions, we describe here a version that is monotonic in $U_2(\theta^\delta)$. Its update is as follows: pick a triplet $(i,j,x)\in P$ and change the variable $\delta_{ij,x}$ such that the equality
\begin{equation}
\theta_{i,x}^\delta = \max_{y\in L} \theta_{ij,xy}^\delta
\label{eq:msd-fp}
\end{equation}
becomes satisfied. Due to~\refeq{reparam}, this is done by setting
\begin{equation}
\delta_{ij,x} \gets \delta_{ij,x} + \tfrac12\bigl( \theta^\delta_{i,x} - \max_{y\in L} \theta^\delta_{ij,xy} \bigr).
\label{eq:msd-update}
\end{equation}
Any point satisfying~\refeq{msd-fp} for all $(i,j,x)\in P$ is a \emph{fixed point} of max-sum diffusion.

\begin{algorithm}
\caption{Max-sum diffusion}
\label{alg:msd}
\begin{algorithmic}
\STATE {\bf Input:} weights $\theta$, initial point $\delta$, precision $\varepsilon \ge 0$
\STATE $\eta \gets \infty$
\WHILE{$\eta \geq \varepsilon$}
\STATE $\eta \gets 0$
\FOR{$(i,j,x) \in P$}
\STATE $d \gets \tfrac12\bigl( \theta^\delta_{i,x} - \max\limits_{y\in L} \theta^\delta_{ij,xy} \bigr)$ 
\STATE $\delta_{ij,x} \gets \delta_{ij,x} + d$
\STATE $\eta \gets \max\{\abs d, \eta\}$
\ENDFOR
\ENDWHILE
\end{algorithmic}
\end{algorithm}

Max-sum diffusion is a particular case of the method from~\S\ref{sec:aff}. The function $\delta\mapsto U_2(\theta^\delta)$ has the form~\refeq{fun}, being the pointwise maximum of affine functions~\refeq{reparam} with coefficients in $\{-1,0,+1\}$ (thus we have $c=C=1$ in~\refeq{slopes}) and offsets~$\theta$. The max-sum diffusion update on triplet $(i,j,x)\in P$ seeks to minimize the maximum over variable~$\delta_{ij,x}$ of only those affine functions that depend on variable~$\delta_{ij,x}$, i.e., the functions $\theta^\delta_{i,x}$ and $\{\theta^\delta_{ij,xy}\}_{y\in L}$, keeping the other variables fixed.
% After simple manipulations, we obtain that this problem has the unique solution~\refeq{msd-update}.
%(i.e., it minimizes $\max\{\theta^\delta_{ij,x},\max_{y\in L}\theta^\delta_{ij,xy}\}$)
For any $d\in\mathbb R$, we have
\begin{IEEEeqnarray}{rClCl"l}
\label{eq:msd-maxf}
\theta^{\delta+de_{ij,x}}_{i,x} &=& \theta^\delta_{i,x} &-& d \IEEEsubnumstart\\
\theta^{\delta+de_{ij,x}}_{ij,xy} &=& \theta^\delta_{ij,xy} &+& d & \forall y\in L
\end{IEEEeqnarray}
where $e_{ij,x}$ is the $(i,j,x$)-th standard basis vector of $\mathbb R^P$. Minimizing the maximum of functions~\refeq{msd-maxf} over~$d$ yields $d=\tfrac12\bigl( \theta^\delta_{i,x} - \max_{y\in L} \theta^\delta_{ij,xy} \bigr)$.
% \begin{equation}
% \max\{ \theta^\delta_{i,x} - d, \; \max_{y\in L} (\theta^\delta_{ij,xy} + d) \}
% \end{equation}

% \tw{Tohle propositoin bych nahradil textem nahore:}

% \begin{proposition}
% Every inner loop of Max-Sum diffusion from Algorithm~\ref{alg:msd} is minimization of maxima of affine functions.
% \end{proposition}
% \begin{proof}
%     The affine functions are (overloading variable $\delta$)
%     \begin{align*}
%         &\theta_{i,x} - \delta,  \\
%         &\theta_{ij,xy} + \delta, \quad \forall y \in L.
%     \end{align*}
%     The minimum of their maxima is clearly attained for such $\delta$ that 
%     \[
%     \theta_{i,x} - \delta = \max_{y\in L} \theta_{ij,xy} + \delta,
%     \]
%     or equivalently, when 
%     \[
%     \delta = \frac{1}{2}\left(\theta_{i,x} - \max_{y\in L} \theta_{ij,xy} \right),
%     \]
%     recovering the update of the algorithm.
% \end{proof}

To meet the assumptions of Theorem~\ref{thm:main_aff}, we show that the weights remain bounded during max-sum diffusion.

\begin{lemma}\label{lem:msd_bound}
Every element of the weight vector $\theta^\delta$ is bounded from below during Algorithm~\ref{alg:msd}. 
\end{lemma}

\vspace*{-2ex}
\begin{proof}
Consider the expression 
\[
\sum_{(i,j,x) \in P} \Bigl( |L|\theta^\delta_{i,x} + \sum_{y \in L}\theta^\delta_{ij,xy} \Bigr) .
\]
It does not depend on any $\delta_{ij,x}$ as they cancel out. Moreover, it is a non-negative combination of the elements of~$\theta^\delta$. Since the maximum  $U_2(\theta^\delta)$ of these elements is bounded from above, each element of~$\theta^\delta$ has to be bounded also from below.
\end{proof}

Now we are ready to state the convergence result:

\begin{theorem}
For any $\varepsilon>0$, Algorithm~\ref{alg:msd} terminates within $\mathcal{O}\left(1/\varepsilon\right)$ steps.
For $\varepsilon=0$, vectors~$\delta$  converge to a max-sum diffusion fixed point, given by~\refeq{msd-fp}.
\end{theorem}

\vspace*{-2ex}
\begin{proof}
Consequence of Lemma~\ref{lem:msd_bound} and Theorem~\ref{thm:main_aff}.
\end{proof}

\subsection{Max-marginal Averaging}

A powerful approach to construct convex bounds on combinatorial optimization problems is Lagrangian decomposition \cite{Guignard-MathProg-1987}. For MAP inference, it was applied by \cite{Johnson-allerton2007} and popularized by \cite{Komodakis-PAMI-2011}. It also underlies the tree-decomposition methods such as TRW-S \cite{Kolmogorov06}.

Many combinatorial optimization problems can be written in the general form
\begin{equation}
F(\theta) = \max_{\mu\in M} \<\theta,\mu\>
\label{eq:Fdecomp}
\end{equation}
where $\theta\in\mathbb R^I$ are weights, $M\subseteq \{0,1\}^I$ is the feasible set (combinatorially large), $I$~is a finite set of `features', and $\<\cdot,\cdot\>$ denotes the dot product.
The MAP inference problem~\refeq{maxsum} is obtained as a special case for $I$ given by~\refeq{I}, $M=\phi(L^V)=\set{\phi(x)\mid x\in L^V}$, and the `feature map' $\phi\colon L^V\to\{0,1\}^I$ being such that the function $\<\theta,\phi(x)\>$ coincides with the objective function of~\refeq{maxsum}. The convex hull of~$M$ is known as the \emph{marginal polytope} \cite{Wainwright08}.

An upper bound on~\refeq{Fdecomp} is constructed by decomposition to
subproblems\footnote{To see how the bound~\refeq{UBdecomp} arises from Lagrangian decomposition, write problem~\refeq{Fdecomp} as minimization of $\sum_{s\in S}\<\theta_s,\mu_s\>$ subject to $\mu_s=\mu$ and $\mu,\mu_s\in M$, and then dualize the coupling constraints $\mu_s=\mu$.}. Let $S$ denote the set of subproblems and $\theta_s\in\mathbb R^I$ the weight vector of subproblem $s\in S$. These weights satisfy
\begin{subequations}
\label{eq:theta-cons}
\begin{align}
\sum_{s\in S}\theta_s &= \theta , \label{eq:theta-cons:1}\\
\theta_{s,i} &= 0 \qquad \forall s\in S, \; i\in I\setminus I_s , \label{eq:theta-cons:2}
\end{align}
\end{subequations}
where $\theta_{s,i}$ denotes the $i$th component of vector~$\theta_s$ and each set $I_s\subseteq I$ is such that the function $F(\theta_s)$ is tractable to compute (e.g., each $I_s$ defines a acyclic subproblem). 
By swapping max and sum in~\refeq{Fdecomp}, we obtain two upper bounds (analogically to~\refeq{F<U<U})
\begin{equation}
F(\theta)
= F\Bigl( \sum_{s\in S} \theta_s \Bigr)
\le \sum_{s\in S} F(\theta_s)
\le |S|\max_{s\in S} F(\theta_s) .
\label{eq:UBdecomp}
\end{equation}
We want to minimize one of the upper bounds in~\refeq{UBdecomp} over the variables $(\theta_s)_{s\in S}$ subject to~\refeq{theta-cons}.

For $I$ and $\phi$ defined by~\refeq{maxsum} and natural choices of
sets $I_s$ (e.g., the rows and columns of an image),
% Consider the undirected graph $(S,J)$ where $\{s,t\}\in J$ iff
% $I_s\cap I_t\neq\emptyset$. If this graph is connected (which is
% natural for reasonable subproblem collections), then by adding
% suitable constants to suitable components of~$\theta_{s,i}$
the numbers $F(\theta_s)$ can always be made equal for all $s\in S$
while keeping~\refeq{theta-cons}. Hence the two upper
bounds in~\refeq{UBdecomp} coincide at optimum. Thus, we can focus only on the second bound.

Minimization of the upper bound in Lagrangian decomposition is traditionally done by subgradient methods, which for the MAP inference problem was applied by \cite{Komodakis-PAMI-2011}. An alternative but less common approach is so-called \emph{max-marginal averaging}, for MAP inference first proposed by \cite{Kolmogorov06,Johnson-allerton2007}. It can be seen as a block-coordinate descent, where in each iteration we pick some $i\in I$ and minimize the upper bound over the variables $(\theta_{s,i})_{s\in S}$ subject to~\refeq{theta-cons}.
We shall describe here its slightly different version (with similar convergent properties), in which only a \emph{pair} of max-marginals is averaged in each update.

%Before we describe it, we reformulate upper bound minimization as an instance of unconstrained minimization of maximum of affine functions, which will allow us to apply our results from~\S\ref{sec:aff}.

% In each update of max-marginal averaging, we pick a feature $i\in I$ and change the variables
% $(\theta_{s,i})_{s\in S_i}$ so that the max-marginals $F_i(\theta_s)$
% become equal for all $s\in S_i$.
% Due to~\refeq{Fi+delta}, this has the unique solution: set $\theta_{s,i}\gets\theta_{s,i}+\delta_s$ where
% \begin{equation}
% \delta_s = |S_i|^{-1}\sum_{t\in S_i} F_i(\theta_t) - F_i(\theta_s) .
% \label{eq:maxmarg-ave-S}
% \end{equation}

The \emph{max-marginal} of the function $\<\theta,\mu\>$ associated with a feature $i\in I$ is the number
\begin{equation}
F_i(\theta)
= \max_{\mu\in M:\,\mu_i=1} \<\theta,\mu\> .
%= \theta_i + \max_{x\in X} \sum_{j\neq i} \theta_j\phi_j(x) .
\label{eq:Fi}
\end{equation}
Note that $F_i$ depends on $\theta_i$ linearly: for any $d\in\mathbb R$ we have
\begin{equation}
F_i(\theta+de_i)=F_i(\theta)+d
\label{eq:Fi+delta}
\end{equation}
where $e_i\in\mathbb R^I$ denotes the $i$th standard basis vector of~$\mathbb R^I$.

Suppose that for each feature $i\in I$ we have chosen a set $E_i\subseteq S_i\times S_i$ where $S_i=\{\,s\in S\mid i\in I_s\,\}$.
One update of pairwise max-marginal averaging proceeds as follows: pick some $i\in I$ and $(s,t)\in E_i$, and change the variables~$\theta_{s,i}$ and~$\theta_{t,i}$ to enforce $F_i(\theta_s)=F_i(\theta_t)$. To maintain~\refeq{theta-cons:1}, due to~\refeq{Fi+delta} this can be done by adding a number $d$ to~$\theta_{s,i}$ and subtracting the same number from~$\theta_{t,i}$. Clearly, such number is uniquely given by\footnote{Computing the involved max-marginals from scratch before every update would be costly in practice. However, it is often possible to reuse partial results from computing earlier max-marginals, which can increase efficiency significantly. Compare, e.g., the algorithms in Figures~1 and~2 in \cite{Kolmogorov06}.}
\begin{equation}
d = \tfrac12 \bigl( F_i(\theta_t) - F_i(\theta_s) \bigr) .
\label{eq:maxmarg-ave}
\end{equation}
This update in fact minimizes $\max\{F_i(\theta_s),F_i(\theta_t)\}$ over $\theta_{s,i}$ and~$\theta_{t,i}$ subject to~\refeq{theta-cons}.

% For that, we parameterize the affine subspace defined by the constraints~\refeq{theta-cons}. Let $(\theta_s)_{s\in S}$ be fixed vectors satisfying \refeq{theta-cons}. For each $i\in I$, denote $S_i=\{\,s\in S\mid i\in I_s\,\}$ and let $E_i\subseteq S_i\times S_i$ be such that the digraph $(S_i,E_i)$ is connected and its edges cover~$S_i$ (the edge orientations do not matter). Any vector satisfying \refeq{theta-cons} has the entries

To apply our results from~\S\ref{sec:aff}, we need to reformulate the upper bound minimization as an \emph{unconstrained} minimization of the maximum of affine functions. For that, it suffices to parameterize the affine subspace defined by~\refeq{theta-cons}. If a vector $(\theta_s)_{s\in S}$ satisfies~\refeq{theta-cons}, then the vector $(\theta^\delta_s)_{s\in S}$ given by
\begin{equation}
\theta_{s,i}^\delta = \theta_{s,i} + \sum_{t\mid(s,t)\in S_i} \delta_{st,i} - \sum_{t\mid(t,s)\in S_i} \delta_{ts,i}
\label{eq:maxmarg-reparam}
\end{equation}
also satisfies~\refeq{theta-cons} for any `messages' $\delta_{st,i}\in\mathbb R$.
%All messages $\delta_{st,i}$ for $i\in I$ and $(s,t)\in E_i$ form a vector~$\delta$.
For instance, if $S_i=\{1,2,3,5\}$ and $E_i=\{(1,2),(2,3),(3,5)\}$, then \refeq{maxmarg-reparam} reads
\begin{IEEEeqnarray*}{rClClClCl}
\theta_{1,i}^\delta &=& \theta_{1,i} &+& \delta_{12,i} \\
\theta_{2,i}^\delta &=& \theta_{2,i} &-& \delta_{12,i} &+& \delta_{23,i} \\
\theta_{3,i}^\delta &=& \theta_{3,i} && &-& \delta_{23,i} &+& \delta_{35,i} \\
\theta_{5,i}^\delta &=& \theta_{5,i} && && &-& \delta_{35,i}
\end{IEEEeqnarray*}
% \[
% \begin{tabular}{LLLL}
% \theta_{1,i}^\delta = \theta_{1,i}
% &+\delta_{12,i} \\
% \theta_{2,i}^\delta = \theta_{2,i} &-\delta_{12,i} & +\delta_{23,i} \\
% \theta_{3,i}^\delta = \theta_{3,i}&&-\delta_{23,i}&+\delta_{35,i} \\
% \theta_{5,i}^\delta = \theta_{5,i}&&&-\delta_{35,i}
% \end{tabular}
% \]
Clearly $\sum_{s\in S_i}\theta_{s,i}^\delta=\sum_{s\in S_i}\theta_{s,i}$ because $\delta_{st,i}$ cancel out.
Note, \refeq{maxmarg-reparam} is a counterpart of~\refeq{reparam}.
If the digraph $(S_i,E_i)$ is connected and its edges~$E_i$ cover~$S_i$, then \emph{any} vector $(\theta^\delta_s)_{s\in S}$ satisfying~\refeq{theta-cons} can be parameterized as~\refeq{maxmarg-reparam}.

\begin{algorithm}
\caption{Max-marginal averaging}\label{alg:maxmarg-ave}
\begin{algorithmic}
\STATE {\bf Input:} weights $(\theta_s)_{s\in S}$, initial point~$\delta$, precision $\varepsilon \ge 0$
\STATE $\eta \gets \infty$
\WHILE{$\eta \geq \varepsilon$}
\STATE $\eta \gets 0$
\FOR{$i\in I$}
\FOR{$(s,t)\in E_i$}
\STATE $d \gets \tfrac12 \bigl( F_i(\theta^\delta_t) - F_i(\theta^\delta_s) \bigr)$
\STATE $\delta_{st,i} \gets \delta_{st,i} + d$
\STATE $\eta \gets \max\{|d|,\eta\}$
\ENDFOR
\ENDFOR
\ENDWHILE
\end{algorithmic}
\end{algorithm}

Now, the upper bound minimization reads as unconstrained minimization of the function $\max_{s\in S} F(\theta^\delta_s)$ over~$\delta$ and pairwise max-marginal averaging becomes Algorithm~\ref{alg:maxmarg-ave}.
The fixed point of the method is any point satisfying $F_i(\theta^\delta_s)=F_i(\theta^\delta_t)$ for all $i\in I$ and $(s,t)\in S_i$.

This is a particular case of the method from~\S\ref{sec:aff}:
the objective function has the form~\refeq{fun}, where the affine functions $x\mapsto a_i^Tx+b_i$ correspond to functions $\delta\mapsto\<\theta^\delta_s,\phi(x)\>$. 
One inner iteration minimizes the maximum of only those affine functions that depend on variable~$\delta_{st,i}$.
Since $\mu$ has non-negative components, it follows from~\refeq{maxmarg-reparam} that whenever a message~$\delta_{st,i}$ changes, some of these functions increase and some decrease, which verifies condition~\refeq{N-scons}. 

\comment{
\begin{lemma}
    The set of values of  $\inner{\theta^s, \phi(x)}$  for any $x \in L^V$ and any $\theta^s \in \{\theta^t\}_{t=1}^S$ attained during the run of Algorithm~\ref{alg:maxmarg-ave} is bounded from below.
\end{lemma}

\vspace*{-2ex}
\begin{proof}
     Note that $\max_{s\in S}F(\theta^s)$ is never increased so we have that $\inner{\theta^s, \phi(x)}$ is bounded from above. Seeking a contradiction, assume that for some $x$ and $s$ we have that $\inner{\theta^s, \phi(x)}$ is not bounded from below throughout the run of the algorithm. Consider now the following identity $\inner{\theta, \phi(x)} = \sum_{s \in S} \inner{\theta^s, \phi(x)}$, where the term on the LHS is a constant and the term on the RHS cannot be bounded from below which is absurd.
\end{proof}
}
\begin{lemma}\label{lem:mma_bound}
The numbers $\inner{\theta^s,\mu}$ for any $\mu\in M$ and $s\in S$ are bounded from below during Algorithm~\ref{alg:maxmarg-ave}.
\end{lemma}

\vspace*{-2ex}
\begin{proof}
As the objective $\max_{s\in S}F(\theta^s)$ never increases during the algorithm, the numbers $\inner{\theta^s,\mu}$ are bounded above. Hence, due to the identity $\inner{\theta,\mu} = \sum_{s \in S} \inner{\theta^s,\mu}$, they are also bounded below because the LHS is constant.
\end{proof}

%Now we are ready to state the main result of this section, the convergence of the max-marginal averaging algorithm.

\begin{theorem}
For any $\varepsilon>0$, Algorithm~\ref{alg:maxmarg-ave} terminates in $\mathcal{O}\left(1/\varepsilon\right)$ steps.
For $\varepsilon =0$, the vectors $\theta_s^\delta$ converge to a fixed point of max-marginal averaging.
\end{theorem}

\vspace*{-2ex}
\begin{proof}
    Consequence of Lemma~\ref{lem:mma_bound} and Theorem~\ref{thm:main_aff}
\end{proof}

% \tw{Again, say also that for $\varepsilon=0$, it converges to a fixed point given by the fixed point condition $F_i(\theta^\delta_s)=F_i(\theta^\delta_t)$ for all $i\in I$ and $(s,t)\in S_i$.}

\section{Mid-point Rule in Coordinate Descent}
\label{sec:midpoint}

When applying BCD to any (possibly non-smooth and/or constrained) convex problem, \cite{Werner-CVPR20} proposed that if block-minimizers are non-unique we should always choose a minimizer from the relative interior of the block-minimizer set. They proved that this \emph{relative interior rule} is not worse, in a precise sense, than any other rule to choose block-minimizers.

The rule we proposed in~\S\ref{sec:aff} for unconstrained minimization of function~\refeq{fun}, namely to ignore the affine functions that do not depend on the current variable, is a special case of the relative interior rule. Indeed, the chosen minimizer of the univariate function $x_j\mapsto f(x)$ is always in the interior of the set of minimizers (which is an interval or a singleton), see Example~\ref{ex:nonunique}.
However, we could choose any other point inside the interval. A natural way is to choose the middle point of the interval. We call this the \emph{mid-point rule}.
For unconstrained minimization of a function~\refeq{fun} with coefficients $A\in\{-1,0,+1\}^{m\times n}$ these two rules clearly coincide (as in Example~\ref{ex:nonunique}), but for more general coefficients they do not.\footnote{Interestingly, the `middle-point algorithm' proposed by \cite{Wedelin-AOR-1995} for dual LP relaxation of the set cover problem is essentially the same as our algorithm from~\S\ref{sec:aff} for this particular problem. He further argued \cite{Wedelin-preprint-2013} that when applied to MAP inference, this algorithm is equivalent to max-sum diffusion. In both cases, the coefficeent are indeed in $\{-1,0,+1\}$.} 
Moreover, unlike the rule from~\S\ref{sec:aff}, the mid-point rule is applicable (assuming that the coordinate minimizer sets are closed intervals or singletons) even for constrained convex problems, such as linear programs.

Further in this section, we show that coordinate descent with the mid-point rule can cycle when applied to a problem other than unconstrained minimization of function~\refeq{fun} with coefficients in $\{-1,0,+1\}$.

\begin{proposition}
Coordinate descent with the mid-point rule can cycle.
\end{proposition}

\vspace*{-2ex}
\begin{proof}
We start with a constrained problem.
Let $M\subseteq\mathbb R^3$ denote the set formed by the 12~points in the first two columns of the following table:
\begin{equation}
\begin{array}{ll@{\hskip3ex}|@{\hskip3ex}l} 
    (-2,0,0) &(2,0,0) & (\underline0,0,0) \\
    (0, -1, 0) &(0,3,0) &(0,\underline1,0) \\
    (0,1,-1) &(0,1,3)  & (0,1,\underline1) \\
    (-1,1,1) &(3,1,1) & (\underline1,1,1) \\
    (1, -2, 1) &(1,2,1) & (1,\underline0,1) \\
    (1,0,-2) &(1,0,2) & (1,0,\underline0) \\
\end{array}
\label{tab:midpoint_points}
\end{equation}
We will later show that each point from~$M$ is an extremal point of the convex hull of~$M$, denoted by $\conv M$.
Consider the problem of minimizing the constant (e.g., zero) objective on the polytope $\conv M$, and apply to it coordinate descent with the mid-point rule.
Thus, whenever we want to find the mid-point w.r.t.\ a coordinate and the current iterate equals to two points  $x,y\in M$ in the other two coordinates, then the next iterate will be their mid-point, $(x+y)/2$. Now we can see that starting from the point $(0,0,0)$ and updating the coordinates in cyclic order, we obtain the iterates in the third column of the table, where the current coordinate is always underlined. In particular, when we take an iterate from row~$i$, at two positions (over which we are not updating) it is equal to the points from the first two columns in row $i+1$, and the third column of row $i+1$ is the average of the previous two points, indices are $\operatorname{mod} 6$. After six updates, we return the initial point $(0,0,0)$, i.e., the method enters a loop.

Let us show that all points in~$M$ are indeed extreme points of $\conv M$. We do this by presenting a supporting hyperplane of $\conv M$ for each point $x\in M$ that passes only through~$x$; thus, $x$~is not a convex combination of any two points from~$M$. The hyperplanes (actually halfspaces) are as follows:
\begin{equation}
\begin{array}{r@{\;}l@{\hskip3ex}r@{\;}l} 
-x_1 &\leq 2   &   8x_1 -5x_2 -3.5x_3 &\leq 16 \\ 
\hskip-2ex -3.5x_1 -8x_2 -5x_3 &\leq 8   &   x_2 &\leq 3 \\ 
\hskip-2ex -5x_1+3.5x_2-8x_3 &\leq 11.5   &   x_3 &\leq 3 \\
\hskip-2ex -8x_1+5x_2+3.5x_3 &\leq 16.5   &   x_1 &\leq 3\\ 
-x_2 &\leq 2   &   3.5x_1+8x_2+5x_3 &\leq 24.5 \\
-x_3 &\leq 2   &   5x_1 -3.5x_2 +8x_3 &\leq 21
\end{array}
\label{eq:midpoint_hyperplanes}
\end{equation}

Having shown cycling for a constrained problem, we proceed to demonstrate it for an unonstrained minimization of a function in the form~\refeq{fun}. Consider the function
\begin{equation}
f(x) = \max\{0, \, \max_i (Ax-b)_i \}
\label{eq:fun-cycle}
\end{equation}
where $Ax\le b$ denotes the system on linear inequalities~\refeq{midpoint_hyperplanes}.
It is not hard to see that coordinate descent with the mid-point rule applied to unconstrained minimization of function~\refeq{fun-cycle} will produce the same iterates as in table~\eqref{tab:midpoint_points}.
\end{proof}

% To give one more insight, note that unconstrained minimization of the function~\refeq{fun} can be written in the epigraph form
% \begin{equation}
% \min\set{ u \mid a_i^Tx+b_i\le u \;\forall i\in[m], \; x\in\mathbb R^n, \; u\in\mathbb R} .
% \label{eq:epi}
% \end{equation}
% Though these two problems have the same global optima, coordinate descent behaves very differently on them. 
% Let $a_i,b_i$ be such that the inequalities $a_i^Tx+b_i\le0$ are the inequalities~\refeq{midpoint_hyperplanes}. Let the initial vector $x\in\mathbb R^3$ be one of the points in~$M$.
% When we minimize problem~\refeq{epi} over variable~$u$, this variable receives the value of the greatest affine function, i.e., zero.
% When we then minimize cyclically over the components of $x$ using the mid-point rule, coordinate desent will cycle. The intuitive reason is that unlike in~\S\ref{sec:aff}, nothing is monotonically decreasing here.
% In fact, unconstrained minimization of function~\refeq{fun} by coordinate descent is in fact equivalent to applying \emph{block}-coordinate descent to problem~\refeq{epi}, with two-variable blocks $(x_j,u)$, $j\in[n]$.

\section{Conclusion}

In this paper, we first studied coordinate descent method applied to unconstrained minimization of the pointwise maximum of affine functions with sparse coefficients. This function is non-smooth and can have non-unique coordinate-minimizers, hence it is hard for coordinate descent. We considered the version of coordinate descent that resolves this ambiguity by ignoring the affine functions not depending on the current variable, which (under the technical assumption~\refeq{N-scons}) ensures uniqueness of coordinate-minimizers.

As our central result, we proved convergence of this method to its fixed point and, more precisely, showed that the method achieves precision $\varepsilon>0$ in $\mathcal{O}(1/\varepsilon)$ iterations. We did this by designing a novel energy function that strictly decreases with every iteration.

Let us remark that the asymptotic upper bound $\mathcal{O}(1/\varepsilon)$ involves a big constant, which depends exponentially on the problem size. Currently we are not sure if this is only an artifact of our analysis or it is inherent to the method. In experiment, we have never observed an exponential dependence of convergence rate on instance size.

Then we showed that two popular representants of dual coordinate descent algorithms (max-sum diffusion and max-marginal averaging) to minimize LP/Lagrangian upper bound on the MAP inference problem are special cases of the above simple method. This proves the long-standing conjecture that these algorithms converge to a fixed point.

Although we presented the convergence results only for max-sum diffusion and marginal-averaging, we believe the proof straightforwardly extends to a number of other dual \mbox{(block-)}coordinate descent methods to minimize an LP/Lagrangian upper bound on combinatorial optimization problems. An exception is, e.g., SRMP \cite{Kolmogorov-PAMI-2015}, which by design does not converge to a fixed point but only to a locally consistent set.

Finally, we showed that a slightly modified version of coordinate descent, in which the coordinate-wise minimizers are chosen as the midpoints of the intervals, can cycle. This provides a further insight to the behaviour of \mbox{(block-)}coordinate descent on non-smooth and/or constrained problems, which is still relatively poorly understood.

\section*{Acknowledgements}

VV was supported by the DFG Cluster of Excellence “Machine Learning – New Perspectives for Science”, EXC 2064/1, project number 390727645 and is thankful for the support of Open Philanthropy.
TW thanks the CTU institutional support (future fund).

\section*{Impact Statement}
We are not aware of any.
\bibliography{werner}
\bibliographystyle{icml2024}

\end{document}

%% file: macros.tex
\usepackage{tabularx}
\newcolumntype{L}{>{$}l<{$}} % math-mode version of "l" column type

\usepackage{IEEEtrantools}
\def\IEEEsubnumstart{\IEEEyesnumber\IEEEyessubnumber*}

\newcommand{\inner}[1]{\left\langle#1\right\rangle}

\def\R{\mathbb{R}}

\newcommand{\abs}[1]{\left|#1\right|}

\def\conv{\mathop{\rm conv}\nolimits}

\def\sort{\mathop{\rm sort}\nolimits}

\def\conv{\mathop{\rm conv}\nolimits}

\def\argmin{\mathop{\rm argmin}}
\def\Argmax{\mathop{\rm Argmax}}

\def\...{,\ldots,}
\def\refeq#1{{(\ref{eq:#1})}}
\def\<{\langle}
\def\>{\rangle}
\DeclarePairedDelimiterX\set[1]\lbrace\rbrace{\def\mid{\;\delimsize\vert\;}\,#1\,} % Optional parameter: $\set[\big]{ x \mid x > 0 }$ or \Big or \bigg etc.

\usepackage{calc}
\newsavebox\CBox
\newcommand\strike[2][0.5pt]{%
  \ifmmode\sbox\CBox{$#2$}\else\sbox\CBox{#2}\fi%
  \makebox[0pt][l]{\usebox\CBox}%  
  {\color{red}\rule[0.5\ht\CBox-#1/2]{\wd\CBox}{#1}}}

%% file: main.bbl
\begin{thebibliography}{38}
\providecommand{\natexlab}[1]{#1}
\providecommand{\url}[1]{\texttt{#1}}
\expandafter\ifx\csname urlstyle\endcsname\relax
  \providecommand{\doi}[1]{doi: #1}\else
  \providecommand{\doi}{doi: \begingroup \urlstyle{rm}\Url}\fi

\bibitem[Abbas \& Swoboda(2022)Abbas and Swoboda]{Abbas-CVPR-2022}
Abbas, A. and Swoboda, P.
\newblock Fastdog: Fast discrete optimization on gpu.
\newblock In \emph{Conf. on Computer Vision and Pattern Recognition (CVPR)}, pp.\  439--449, 2022.

\bibitem[Bertsekas(1999)]{Bertsekas99}
Bertsekas, D.~P.
\newblock \emph{Nonlinear Programming}.
\newblock Athena Scientific, Belmont, MA, 2nd edition, 1999.

\bibitem[Chen et~al.(2015)Chen, Schwing, Yuille, and Urtasun]{Chen-ICML-2015}
Chen, L.-C., Schwing, A., Yuille, A., and Urtasun, R.
\newblock Learning deep structured models.
\newblock In \emph{Intl. Conf. on Machine Learning (ICML)}, pp.\  1785--1794, 2015.

\bibitem[Cooper et~al.(2010)Cooper, de~Givry, Sanchez, Schiex, Zytnicki, and Werner]{Cooper-AI-2010}
Cooper, M.~C., de~Givry, S., Sanchez, M., Schiex, T., Zytnicki, M., and Werner, T.
\newblock Soft arc consistency revisited.
\newblock \emph{Artificial Intelligence}, 174\penalty0 (7-8):\penalty0 449--478, 2010.

\bibitem[Globerson \& Jaakkola(2008)Globerson and Jaakkola]{Globerson08}
Globerson, A. and Jaakkola, T.
\newblock Fixing max-product: Convergent message passing algorithms for {MAP} {LP}-relaxations.
\newblock In \emph{Neural Information Processing Systems}, pp.\  553--560, 2008.

\bibitem[Guignard \& Kim(1987)Guignard and Kim]{Guignard-MathProg-1987}
Guignard, M. and Kim, S.
\newblock Lagrangean decomposition: {A} model yielding stronger {L}agrangean bounds.
\newblock \emph{Mathematical Programming}, 39:\penalty0 215--228, 1987.

\bibitem[Johnson et~al.(2007)Johnson, Malioutov, and Willsky]{Johnson-allerton2007}
Johnson, J.~K., Malioutov, D.~M., and Willsky, A.~S.
\newblock {L}agrangian relaxation for {MAP} estimation in graphical models.
\newblock In \emph{45th Allerton Conference on Communication, Control and Computing}, 2007.

\bibitem[Kappes et~al.(2015)Kappes, Andres, Hamprecht, Schn{\"o}rr, Nowozin, Batra, Kim, Kausler, Kr{\"o}ger, Lellmann, Komodakis, Savchynskyy, and Rother]{Kappes-study-2015}
Kappes, J.~H., Andres, B., Hamprecht, F.~A., Schn{\"o}rr, C., Nowozin, S., Batra, D., Kim, S., Kausler, B.~X., Kr{\"o}ger, T., Lellmann, J., Komodakis, N., Savchynskyy, B., and Rother, C.
\newblock A comparative study of modern inference techniques for structured discrete energy minimization problems.
\newblock \emph{Intl.\ J.\ of Computer Vision}, 115\penalty0 (2):\penalty0 155--184, 2015.

\bibitem[Kolmogorov(2006)]{Kolmogorov06}
Kolmogorov, V.
\newblock Convergent tree-reweighted message passing for energy minimization.
\newblock \emph{IEEE Trans. Pattern Analysis and Machine Intelligence}, 28\penalty0 (10):\penalty0 1568--1583, 2006.

\bibitem[Kolmogorov(2015)]{Kolmogorov-PAMI-2015}
Kolmogorov, V.
\newblock A new look at reweighted message passing.
\newblock \emph{IEEE Trans.\ on Pattern Analysis and Machine Intelligence}, 37\penalty0 (5), May 2015.

\bibitem[Komodakis et~al.(2011)Komodakis, Paragios, and Tziritas]{Komodakis-PAMI-2011}
Komodakis, N., Paragios, N., and Tziritas, G.
\newblock {MRF} energy minimization and beyond via dual decomposition.
\newblock \emph{IEEE Transactions on Pattern Analysis and Machine Intelligence}, 33\penalty0 (3):\penalty0 531--552, 2011.

\bibitem[Kovalevsky \& Koval(1975)Kovalevsky and Koval]{Kovalevsky-diffusion}
Kovalevsky, V.~A. and Koval, V.~K.
\newblock A diffusion algorithm for decreasing the energy of the max-sum labeling problem.
\newblock {Glushkov Institute of Cybernetics, Kiev, USSR. Unpublished}, 1975.

\bibitem[Lange \& Swoboda(2021)Lange and Swoboda]{Lange-ICML-2021}
Lange, J. and Swoboda, P.
\newblock Efficient message passing for 0-1 {ILPs} with binary decision diagrams.
\newblock In \emph{Intl. Conf. on Machine Learning (ICML)}, volume 139, pp.\  6000--6010. {PMLR}, 2021.

\bibitem[Martins et~al.(2011)Martins, Figueiredo, Aguiar, Smith, and Xing]{Martins:ICL:2011}
Martins, A.~L., Figueiredo, M. A.~T., Aguiar, P. M.~Q., Smith, N.~A., and Xing, E.~P.
\newblock An augmented {L}agrangian approach to constrained {MAP} inference.
\newblock In \emph{Intl.\ Conf.\ on Machine Learning}, pp.\  169--176, 2011.

\bibitem[Meltzer et~al.(2009)Meltzer, Globerson, and Weiss]{Meltzer-UAI-2009}
Meltzer, T., Globerson, A., and Weiss, Y.
\newblock Convergent message passing algorithms: a unifying view.
\newblock In \emph{Conf.\ on Uncertainty in Artificial Intelligence}, pp.\  393--401, 2009.

\bibitem[Meseguer et~al.(2006)Meseguer, Rossi, and Schiex]{Meseguer06}
Meseguer, P., Rossi, F., and Schiex, T.
\newblock Soft constraints.
\newblock In \emph{Handbook of Constraint Programming}, chapter~9. Elsevier, 2006.

\bibitem[Munda et~al.(2017)Munda, Shekhovtsov, Kn{\"o}belreiter, and Pock]{Munda-GCPR-2017}
Munda, G., Shekhovtsov, A., Kn{\"o}belreiter, P., and Pock, T.
\newblock Scalable full flow with learned binary descriptors.
\newblock In \emph{German Conf. on Pattern Recognition (GCPR)}, 2017.

\bibitem[Ruozzi \& Tatikonda(2013)Ruozzi and Tatikonda]{Ruozzi-2013}
Ruozzi, N. and Tatikonda, S.
\newblock Message-passing algorithms: Reparameterizations and splittings.
\newblock \emph{IEEE Transactions on Information Theory}, 59\penalty0 (9):\penalty0 5860--5881, 2013.

\bibitem[Savchynskyy(2012)]{Savchynskyy-CVPR-2012}
Savchynskyy, B.
\newblock A bundle approach to efficient {MAP}-inference by {L}agrangian relaxation.
\newblock In \emph{IEEE Conf.\ on Computer Vision and Pattern Recognition}, pp.\  1688--1695, 2012.

\bibitem[Savchynskyy(2019)]{Bogdan-book-2019}
Savchynskyy, B.
\newblock Discrete graphical models -- an optimization perspective.
\newblock \emph{Foundations and Trends in Computer Graphics and Vision}, 11\penalty0 (3-4):\penalty0 160--429, 2019.
\newblock ISSN 1572-2740.

\bibitem[Savchynskyy et~al.(2012)Savchynskyy, Schmidt, Kappes, and Schn\:orr]{SavchynskyySKS12}
Savchynskyy, B., Schmidt, S., Kappes, J.~H., and Schn\:orr, C.
\newblock Efficient {MRF} energy minimization via adaptive diminishing smoothing.
\newblock In \emph{Conf.\ on Uncertainty in Artificial Intelligence}, pp.\  746--755, 2012.

\bibitem[Schlesinger \& Antoniuk(2011)Schlesinger and Antoniuk]{Schlesinger-2011}
Schlesinger, M.~I. and Antoniuk, K.
\newblock Diffusion algorithms and structural recognition optimization problems.
\newblock \emph{Cybernetics and Systems Analysis}, 47:\penalty0 175--192, 2011.
\newblock ISSN 1060-0396.

\bibitem[Sontag et~al.(2012)Sontag, Globerson, and Jaakkola]{Sontag-MITbook-2012}
Sontag, D., Globerson, A., and Jaakkola, T.
\newblock Introduction to dual decomposition for inference.
\newblock In Sra, S., Nowozin, S., and Wright, S.~J. (eds.), \emph{Optimization for Machine Learning}. MIT Press, 2012.

\bibitem[Swoboda \& Andres(2017)Swoboda and Andres]{Swoboda-CVPR-2017c}
Swoboda, P. and Andres, B.
\newblock A message passing algorithm for the minimum cost multicut problem.
\newblock In \emph{Conf.\ on Computer Vision and Pattern Recognition}, 2017.

\bibitem[Swoboda et~al.(2017{\natexlab{a}})Swoboda, Kuske, and Savchynskyy]{Swoboda-CVPR-2017}
Swoboda, P., Kuske, J., and Savchynskyy, B.
\newblock A dual ascent framework for {L}agrangean decomposition of combinatorial problems.
\newblock In \emph{Conf.\ on Computer Vision and Pattern Recognition}, 2017{\natexlab{a}}.

\bibitem[Swoboda et~al.(2017{\natexlab{b}})Swoboda, Rother, Abu~Alhaija, Kainmuller, and Savchynskyy]{Swoboda-CVPR-2017b}
Swoboda, P., Rother, C., Abu~Alhaija, H., Kainmuller, D., and Savchynskyy, B.
\newblock A study of {L}agrangean decompositions and dual ascent solvers for graph matching.
\newblock In \emph{Conf.\ on Computer Vision and Pattern Recognition}, 2017{\natexlab{b}}.

\bibitem[Szeliski et~al.(2006)Szeliski, Zabih, Scharstein, Veksler, Kolmogorov, Agarwal, Tappen, and Rother]{Szeliski06}
Szeliski, R., Zabih, R., Scharstein, D., Veksler, O., Kolmogorov, V., Agarwal, A., Tappen, M., and Rother, C.
\newblock A comparative study of energy minimization methods for {M}arkov random fields.
\newblock In \emph{Eur.\ Conf.\ on Computer Vision}, pp.\  II: 16--29, 2006.

\bibitem[Tourani et~al.(2018)Tourani, Shekhovtsov, Rother, and Savchynskyy]{Tourani-ECCV-2018}
Tourani, S., Shekhovtsov, A., Rother, C., and Savchynskyy, B.
\newblock Mplp++: Fast, parallel dual block-coordinate ascent for dense graphical models.
\newblock In \emph{The European Conference on Computer Vision (ECCV)}, 2018.

\bibitem[Tourani et~al.(2020)Tourani, Shekhovtsov, Rother, and Savchynskyy]{Tourani-AISTAT-2020}
Tourani, S., Shekhovtsov, A., Rother, C., and Savchynskyy, B.
\newblock Taxonomy of dual block-coordinate ascent methods for discrete energy minimization.
\newblock In \emph{Intl. Conf. on Artificial Intelligence and Statistics}, pp.\  2775--2785, 2020.

\bibitem[Tseng(2001)]{Tseng:2001}
Tseng, P.
\newblock Convergence of a block coordinate descent method for nondifferentiable minimization.
\newblock \emph{J. Optim. Theory Appl.}, 109\penalty0 (3):\penalty0 475--494, June 2001.

\bibitem[Wainwright \& Jordan(2008)Wainwright and Jordan]{Wainwright08}
Wainwright, M.~J. and Jordan, M.~I.
\newblock Graphical models, exponential families, and variational inference.
\newblock \emph{Foundations and Trends in Machine Learning}, 1\penalty0 (1-2):\penalty0 1--305, 2008.

\bibitem[Warga(1963)]{Warga-1963}
Warga, J.
\newblock Minimizing certain convex functions.
\newblock \emph{Journal of the Society for Industrial and Applied Mathematics}, 11\penalty0 (3):\penalty0 588--593, 1963.

\bibitem[Wedelin(1995)]{Wedelin-AOR-1995}
Wedelin, D.
\newblock An algorithm for large scale 0-1 integer programming with application to airline crew scheduling.
\newblock \emph{Ann. Oper. Res.}, 57\penalty0 (1):\penalty0 283--301, 1995.

\bibitem[Wedelin(2013)]{Wedelin-preprint-2013}
Wedelin, D.
\newblock Revisiting the in-the-middle algorithm and heuristic for integer programming and the max-sum problem.
\newblock Chalmers University preprint, \url{https://research.chalmers.se/en/publication/190782}, 2013.

\bibitem[Werner(2007)]{Werner-PAMI07}
Werner, T.
\newblock A linear programming approach to max-sum problem: A review.
\newblock \emph{IEEE Trans. Pattern Analysis and Machine Intelligence}, 29\penalty0 (7):\penalty0 1165--1179, July 2007.

\bibitem[Werner(2010)]{Werner-PAMI-2010}
Werner, T.
\newblock Revisiting the linear programming relaxation approach to {G}ibbs energy minimization and weighted constraint satisfaction.
\newblock \emph{IEEE Trans. Pattern Analysis and Machine Intelligence}, 32\penalty0 (8):\penalty0 1474--1488, August 2010.

\bibitem[Werner(2017)]{Werner-TR-2017-05}
Werner, T.
\newblock On coordinate minimization of piecewise-affine functions.
\newblock Technical Report CTU-CMP-2017-05, Dept.\ of Cybernetics, Fac.\ of Electrical Eng., Czech Technical Univ.\ in Prague, September 2017.

\bibitem[Werner et~al.(2020)Werner, Pr{\r u}{\v s}a, and Dlask]{Werner-CVPR20}
Werner, T., Pr{\r u}{\v s}a, D., and Dlask, T.
\newblock Relative interior rule in block-coordinate descent.
\newblock In \emph{Conf. on Computer Vision and Pattern Recognition (CVPR)}, June 2020.

\end{thebibliography}
